\documentclass[letterpaper,journal]{IEEEtran}

\IEEEoverridecommandlockouts                              

\usepackage{amsmath,amssymb}
\usepackage{graphicx}
\usepackage{algorithmic}
\usepackage{orcidlink}

\let\amssymbboxplus\boxplus
\let\amssymbboxminus\boxminus

\renewcommand{\boxplus}{\mathbin{\mathop\amssymbboxplus}}
\renewcommand{\boxminus}{\mathbin{\mathop\amssymbboxminus}}
 \usepackage{url}

\newcommand{\Real}{\mathbb R}
\usepackage{xcolor}
\newcommand{\red}[1]{{#1}}

\usepackage{orcidlink}
\usepackage{balance}

\title{\LARGE \bf
Constant Time-Delay Leader Following with Neural Networks and Invariant Extended Kalman Filters for Arbitrary Trajectories
}

\author{Luka Antonyshyn$^{1}$, Paulo Ricardo Marques de Araujo$^{2}$, Sidney Givigi$^{2}$,~\IEEEmembership{Senior Member,~IEEE}
\thanks{$^{1}$Luka Antonyshyn is with the University of Toronto Institute for Aerospace Studies, Toronto, Canada
        {\tt\footnotesize luka.antonyshyn@utoronto.ca}.}%
\thanks{$^{2}$Paulo Ricardo Marques de Araujo and Sidney Givigi are with the School of Computing, Queen's University, Kingston, Canada
        {\tt\footnotesize \{paulo.araujo, sidney.givigi\}@queensu.ca}.}%
}

\begin{document}

\maketitle
\thispagestyle{empty}
\pagestyle{empty}

\markboth{IEEE Robotics and Automation Letters}%
{Shell \MakeLowercase{\textit{et al.}}: Bare Demo of IEEEtran.cls for IEEE Journals}
%



\maketitle

\begin{abstract}
This paper proposes a constant time-delay trajectory tracking method for vehicle convoys operating without inter-vehicle communication, a common coordinate system, or global positioning. The method integrates a probabilistic sequence-to-sequence~(Seq2Seq) neural network with an invariant extended Kalman filter~(IEKF) to warm-start the prediction process, allowing accurate estimation of a leader vehicle's relative trajectory on the $\text{SE}(2)$ manifold. A geometric model predictive controller is further incorporated to fully exploit the manifold-based trajectory predictions for improved control performance. The system can handle arbitrary nonlinear trajectories with varying speeds and motion profiles while reducing the need for expert-based domain knowledge for the design of trajectory following systems, even under long trajectory delays. The effectiveness of the method is validated through comparisons with a pure IEKF baseline\red{, learning-based methods, and} the ground-truth trajectory in kinematic simulations, as well as in experiments using real robotic vehicles.
\end{abstract}

\begin{IEEEkeywords}
Autonomous Vehicle Navigation, Integrated Planning and Learning, Data-based approaches (learning, deep learning, reinforcement learning), Connected and Autonomous Vehicles, Multi-agent systems.
\end{IEEEkeywords}

%
\IEEEpeerreviewmaketitle

\section{Introduction}
Autonomous vehicles are increasingly used in transportation, logistics networks, and military applications. They often operate in convoys or platoons for efficiency or group-based tasks~\cite{nahavandi_autonomous_2022}. In civilian settings, convoys often rely on inter-vehicle centralized or decentralized communications for localization and planning. However, to improve scaling to larger number of vehicles and reduce reliance on potentially insecure or unreliable networks, it is of interest to explore non-communicative approaches to this problem.

In numerous military scenarios, a vehicle may need to follow directly in the trajectory of a preceding vehicle to avoid some non-perceptible obstacles or to perform reconnaissance on a given target. This is called the leader-follower convoy problem~\cite{ZhaoConvoy}. A wide range of approaches have been considered for the tracking of a leading vehicle trajectory in such scenarios, including vehicle-to-vehicle communication-based approaches~\cite{ZhaoConvoy, WuConsensus,SantosFuzzy,zhang_leader-follower_2024,han_leader-follower_2024}, and sensor-based relative localization methods~\cite{petrov_mathematical_nodate,petrov_nonlinear_nodate,monkeySee,denasi_independent_2018,walter_uvdar_2019}. 

Additionally, it may be desirable to perform this task without communications, either to reduce detectability or to protect a system from communication jamming or spoofing~\cite{Yaacoub2022}. As such, vehicles must be able to keep track of the desired trajectories using only locally observable information, such as the leading vehicle's speed, relative position, and proprioceptive information. This is sometimes referred to as the no-communication autonomous vehicle following problem~\cite{petrov_mathematical_nodate} or the leader-following problem~\cite{monkeySee}. 

Many approaches to localization and leader trajectory tracking problems assume that vehicles exist in a vector space $\Real^{3}$, utilizing either overparameterizations of the states or using workarounds, such as switching between orderings of parameterizations, to avoid singularities~\cite{HERTZBERG201357}. Vehicle configurations are often expressed more naturally on topological spaces, such as the Special Euclidean group $\text{SE}(2)$. Previous works have shown that using an $\text{SE}(2)$ representation can result in superior tracking~\cite{gmpc} and localization~\cite{iekf}. 

In this work, we propose a \red{system for} no-communication deep learning-based constant time-delay leader-follower trajectory estimation on the $\text{SE}(2)$ manifold. We further integrate this system with a GMPC to improve the tracking error. Our contributions are as follows:
\begin{itemize}
    \item We propose a neural network (NN) architecture based on sequence-to-sequence models to track time-delayed trajectories of a leader vehicle in non-communicating convoys, allowing the system to follow arbitrary nonlinear trajectories.
    \item We detail a process for generating datasets for training a convoy trajectory-following model using simulations.
    \item We propose modifications to Geometric Model Predictive Control (GMPC) for integration with our adaptive trajectory tracking model.
    \item We evaluate our method using kinematic simulations against baseline methods, and validate it with robot experiments.
\end{itemize}

The paper is structured as follows: Section \ref{sec:related} explores related works, Section \ref{sec:background} covers the mathematical and conceptual background, Section \ref{sec:Method} details the proposed system, Section \ref{sec:Sims} presents numerical simulations, \red{ablations} and comparisons, and Section \ref{sec:exp} presents and analyses real-robot experiments.
\section{Related Works} \label{sec:related}

\subsection{Mobile Robot Convoys}

Autonomous convoying involves several aspects, including preceding vehicle pose estimation, leader trajectory management and trajectory tracking~\cite{nahavandi_autonomous_2022}. As we utilize arbitrary leader trajectories, we focus on pose estimation and trajectory tracking in this review.

In many scenarios, vehicle-to-vehicle communication is a viable method for sharing the information necessary to track a leader's trajectory. Santos et al.~\cite{SantosFuzzy} propose a fuzzy controller for a decentralized overlapping system in which vehicles communicate with the preceding and following robot to select setpoints in nonlinear trajectories. Wu et al.~\cite{WuConsensus} develop a consensus-based control framework to design a leader-follower convoy formation controller. Zhao et al.~\cite{ZhaoConvoy} integrate computer vision, an extended Kalman filter (EKF), waypoint management, and distance control for leader-follower tracking. Han et al.~\cite{han_leader-follower_2024} design a waypoint-based formation controller to stabilize the follower vehicle behaviour after passing a waypoint set by the leader. Zhang et al.~\cite{zhang_leader-follower_2024} address nonholonomic agents tracking parametric curves with a formation controller using Lyapunov stability analysis and directed communication.

When inter-vehicle communication is not viable, follower vehicles must track the relative poses of the leader vehicle and their own motion to reconstruct the intended trajectory. In \cite{cheung}, Cheung et al. use behaviour-based methods to ensure tracking in the case of communication jamming through \red{prior waypoint information}. 

Computer vision-based methods for estimating the relative position and movement of a leader vehicle have also been proposed and fall into template-based and model-based. Fries et al.~\cite{Fries} use a template-based computer vision method to initialize model-based particle filters upon tracking loss. Wang et al. \cite{WangVision} combine template-based computer vision with low-power WiFi for trajectory tracking in GPS-denied environments. Denasi et al.~\cite{denasi_independent_2018} apply image processing for fixed-frame localization of a leader and follower micro-robot. Walter et al.~\cite{walter_uvdar_2019} employ high-dimensional sensor data in the form of UVDAR and real-time trajectory planning to adapt behaviours based on leader distance. \red{These works tend to focus on the perception of the leader vehicle, thus either discarding the problem of trajectory tracking or assuming a predefined structure in the problem.}

Non-vision based approaches to estimating the position of a leader vehicle in non-communicative scenarios have also been explored. Petrov~\cite{petrov_mathematical_nodate,petrov_nonlinear_nodate} derives kinematic equations for a leader-follower convoy problem and develops an adaptive tracking controller to handle unknown leader velocities. Similarly, Wang et al.~\cite{monkeySee} use a pairwise Kalman filter to predict leader waypoints with a linear model predictive longitudinal controller that inversely weighs the uncertainty of estimated waypoints. They implement a proportional controller to correct follower orientation and reduce tracking errors. 

\subsection{Neural Network-Based Trajectory Reconstruction}

Several methods reliant on recurrent NNs have been proposed for trajectory reconstruction and prediction from sensor observations\red{, primarily for single-vehicle settings}. Shrivastava et al.\cite{shrivastava_deep_2021} use geographically-based clustering, nearest neighbour searches and LSTM networks to reconstruct \red{single} vehicle trajectories using partial information. Sun et al.~\cite{sun2021idol} use LSTM networks combined with a extended Kalman filter to predict a phones position and orientation on the $\text{SE}(2)$ manifold using inertial measurement unit readings. Zamboni et al.~\cite{ZAMBONI2022108252} instead use convolutional NNs to predict pedestrian trajectories. Their approach uses data augmentation, and achieves state of the art performance.

Using sequence-to-sequence networks for vehicle trajectory prediction and reconstruction has also been explored. Park et al.~\cite{Park2018} utilize an LSTM-based Seq2Seq architecture to generate relative occupancy grid maps for multi-vehicle scenarios. Zhang et al.~\cite{Zhang2013A} consider trajectory prediction on a ship. They utilize a Seq2Seq network architecture, using information on the ships movement to guide the decoder.

\red{These works focus on the reconstruction of ego-vehicle trajectories.} To our knowledge, no methods have explored $\text{SE}(2)$-based trajectory estimation on a manifold, or deep-learning for leader trajectory prediction in a no-communication, time-delay leader-following setting. By combining these approaches, a GMPC-based system can be designed that uses trajectory information to improve tracking performance while reducing reliance on domain expert knowledge. \red{In our method, we do not directly consider the use of perception models for the explicit observation of the lead vehicle or dynamic obstacle avoidance, instead electing to utilize an abstraction of an observation. This allows us to focus on the development of the core trajectory estimation and control components of the system.}
\section{Preliminaries}\label{sec:background}

\subsection{Special Euclidean Groups}
A manifold is a topology that locally resembles Euclidean space. In control and robotics, manifolds naturally represent configuration spaces. For planar autonomous vehicles, the configuration exists on the $\text{SE}(2)$ group. The configuration $\chi \in \Real^{3\times 3}$ of a rigid body in $\text{SE}(2)$ can be represented by a homogeneous transformation matrix
\begin{align}
    \chi = \begin{bmatrix}
        \cos(\theta) & -\sin(\theta) & x \\
        \sin(\theta) & \cos(\theta) & y \\
        0     &       0      &     1
    \end{bmatrix} \in \Real^{3\times 3}
\end{align}
with $x,y$ referring to the position of the body in a $\text{2D}$ plane and $\theta \in [0, 2\pi]$ referring to the orientation of the body.

Manifolds are curved spaces where traditional addition and subtraction operations are not valid.
Hertzberg et al.~\cite{HERTZBERG201357} introduce the boxminus operator, $\boxminus$, to account for these differences in orientations on a manifold. 

For the Special Orthogonal group $\text{SO}(2)$ that represents orientations in $\text{2D}$, this operator is defined as
\begin{eqnarray*}
    \alpha \boxminus \beta &=& v_{\pi}(\alpha - \beta) \\
    \text{with } v_{\pi}(\delta) &=& \delta - 2\pi \frac{\delta + \pi}{2\pi}
\end{eqnarray*}
which defines $\alpha \boxminus \beta$ as the smallest change in angle needed to rotate $\beta$ into $\alpha$. For the $\text{SE}(2)$ space, we may take the Cartesian of the vector space $\Real^{2}$ and $\text{SO}(2)$ and define the $\boxminus$ operation component wise as
\begin{align}
    \begin{bmatrix}
        x_{1} & x_{2}
    \end{bmatrix} \boxminus \begin{bmatrix}
        \delta_{1} & \delta_{2}
    \end{bmatrix}  = \begin{bmatrix}
        x_{1} \boxminus \delta_{1} & x_{2} \boxminus \delta_{2}
    \end{bmatrix}
\end{align}
with $\boxminus$ for the $\Real^{2}$ space being vector subtraction.

\subsection{Geometric Model Predictive Control}

In \cite{gmpc}, Tang et al. consider a problem of a unicycle following a trajectory on an $\text{SE}(2)$ manifold. By utilizing a more natural representation of the configuration of a robot, they improve tracking performance when compared to approaches that optimize over a vector space.

We may state the error between a configuration and a desired trajectory as
\begin{align}
  \Psi(t) = \chi_{d}(t)^{-1}\chi(t) \in \text{SE}(2)
\end{align}
with $\chi_{d}(t)$ being the desired configuration at time $t$.

Following this, we are able to derive the error dynamics on the manifold and state the problem as a continuous-time nonlinear optimization problem:
\begin{align*}
    \min_{\mathbf{u}(t)} \phi\big(\Psi(T)\big) + \int_{0}^{T} l(\Psi{\tau}, \mathbf{u}(\tau))d\tau \\
    s.t. \dot{\Psi}{\tau} = \Psi(\tau) \boldsymbol{\zeta}(t)^{\wedge}
    - \boldsymbol{\zeta}_d(t) \Psi(t), \\
    \boldsymbol{\zeta}(t) = C(0)\mathbf{u}(t), \\
    u_{\text{min}} \leq \mathbf{u}(t) \leq u_{\text{max}}, \\
    \Psi(0) = \Psi_{\text{init}}
\end{align*}
where $\phi$ refers to a terminal cost, $l$ refers to a running cost, $C(0)$ is a constant matrix, $T$ is a terminal time, $u_{\text{min}}$ and $u_{\text{max}}$ refer to lower and upper bounds on the inputs, and $\Psi_{\text{init}}$ is the initial error. Using an approximation, the problem is reformulated as a discrete-time convex optimization problem:
\begin{align*}
    \boldsymbol{\psi}(T)^{T} Q_{f} \boldsymbol{\psi}(T) + \sum_{t=0}^{T-1} \boldsymbol{\psi}(t)^{T} Q \boldsymbol{\psi}(t) + \mathbf{u}_{k}^{T} H_{k} \mathbf{u}_{k} \\
    s.t. \boldsymbol{\psi}_{t+1} = A_{t}\boldsymbol{\psi}_{t} + B_{t}\mathbf{u}_{t} + \mathbf{c}_{t}, \\
    u_{\text{min}} \leq \mathbf{u}(t) \leq u_{\text{max}}, \\
    \boldsymbol{\psi}(0) = \boldsymbol{\psi}_{\text{init}}
\end{align*}
where $Q,H,A_{t}, B_{t}, c_{t}$ can be obtained using numerical integration, $\boldsymbol{\psi}$ refers to the error, and $Q_{f}, Q, H$ are the terminal, running state and running input costs respectively. We omit much of the derivation and details for brevity. A more rigorous treatment of the problem can be found in \cite{gmpc}.

\subsection{Invariant Extended Kalman Filters}\label{sec:IEKF}
A filter is an observer method that combines a model of a system's dynamics and measurements of the system to estimate the true system states. 
When considering non-linear systems, a common approach is to linearize the system about the current estimated state, resulting in the extended Kalman filter~(EKF). While adequate in many cases, when the estimate is far from the true state of the system, the EKF estimate can diverge~\cite{iekf}.

Barrau and Bonnabel~\cite{iekf} consider a general nonlinear system, represented on the $\text{SE}(2)$ group. We represent the measurement model of the system as
\begin{align}
    Y_{t} = h(\chi_t) + V_t
\end{align}
with $h$ being an observation function that maps states to system outputs and $V_t$ being unknown measurement noise at time $t$. Additionally, we state the (left) invariant error, used henceforth in this work, between a predicted and measured state on the $\text{SE}(2)$ group as
\begin{align}
    \eta = \chi^{-1}\hat{\chi}
\end{align}
with $\hat{\chi}$ being the estimated state. This error can be shown to be independent of the trajectory for a large class of systems, allowing the IEKF to provide stronger local convergence guarantees~\cite{iekf}. Additionally, it directly provides estimates on the $\text{SE}(2)$ group, which simplifies integration with geometric control methods.

The IEKF is defined through two steps. The first is the propagation step, where the previous posterior estimate $\hat{\chi}^{+}_{t-1}$ of the state and the previous input $\mathbf{u}_{t-1}$ are used to predict the current state prior. This step is
\begin{align}
    \frac{d}{dt}\hat{\chi}_{t}^{-} = \hat{\chi}_{t-1}^{+}\mathbf{u}_{t-1}^{\wedge} \\
    \Gamma^- = F_t \Gamma_{t-1}^{+} F_t^{-1} + N_t
\end{align}
where $\hat{\chi}_{t}^{-}$ refers to the prior estimate of the state at time $t$, $\Gamma_t$ refers to the covariance matrix, $N_t$ is the covariance of the process noise, and $F_t$ is a matrix representation of
\begin{align}
    g(\gamma^{-1}  \chi, \mathbf{u}) = f(\gamma,\mathbf{u})^{-1}f(\chi,\mathbf{u})
\end{align}
for each $\mathbf{u}$ such that $g(\exp(\zeta),\mathbf{u}) = \text{exp}(F\zeta)~\forall \chi, \gamma \in \text{SE(2)}, \zeta \in \mathfrak{se}(2)~\text{and}~\mathbf{u} \in \mathbb{R}^j$ with $j$ representing the number of inputs of the system.
Before proceeding to the update step, we first define a first order Taylor expansion of the measurement noise
\begin{align*}
    h(\hat{\chi}_{t}^{-}\text{exp}(\xi)) - \chi_{t}^{-} =H_t (\xi) + O(||\xi||^{2})
\end{align*}
with $O$ denoting the order, and $\xi$ denoting some  in the Lie algebra. Following the calculation of the prior, we perform the update step, where the current observation of the system $Y_t$ is used to transform the prior $\hat{\chi}_{t}^{-}$ into the posterior $\hat{Q}_{t}^{+}$. These are:
\begin{align}
        z_t = Y_t - h(\hat{\chi}^{-}_t), \\
        S_t = H_t P_{t}^{-} H_t^T + R_t \\
        K_t = P_t^{-} H^{T}_t S_t^{-1} \\
        \Gamma_t^+ = (I-K_t H_t)\Gamma_t^{-} \\ 
        \hat{\chi}_{t}^{+} = \hat{\chi}_{t}^{-}\text{exp}(K_t z_t) 
\end{align}
with $z_t$ being the innovation, $K_t$ being the gain, and $\hat{\chi}_{t}^{+}$ being the posterior estimate of the state at time $t$. Further discussion on the derivation and theoretical properties of the IEKF are explored in \cite{iekf}.

\subsection{Sequence-to-Sequence Models}
Sequence-to-sequence, or Seq2Seq, models are an architecture of recurrent neural networks with applications in a large range of natural language processing tasks, such as machine translation and image caption generation, among others~\cite{seq2seqsurv}. They consist of two components, an encoder layer that takes the inputs and outputs the networks hidden values and output values, and a decoder layer, that takes the hidden values of the encoder layer and calculates the overall architecture outputs.

Seq2Seq models have also been applied to numerous time-series forecasting problems, such as anomaly detection and financial market prediction, due to their ability to capture complex nonlinear temporal dynamics~\cite{seq2seqfore}. These characteristics also make Seq2Seq models suitable for forecasting and trajectory tracking in dynamic systems.

\section{Proposed Method}\label{sec:Method}

\subsection{Problem Description}

The leader-following problem involves controlling a follower vehicle to track a trajectory defined by the \red{traversed} path of a leader vehicle~\cite{nahavandi_autonomous_2022}. We consider a non-communicating convoy with one following vehicle. \figurename~\ref{fig:leader-follower} presents a graphical depiction of the problem.

\newdimen\cxl
\newdimen\cyl

\newdimen\cxlf
\newdimen\cylf

\newdimen\cxls
\newdimen\cyls

\newdimen\cxlt
\newdimen\cylt

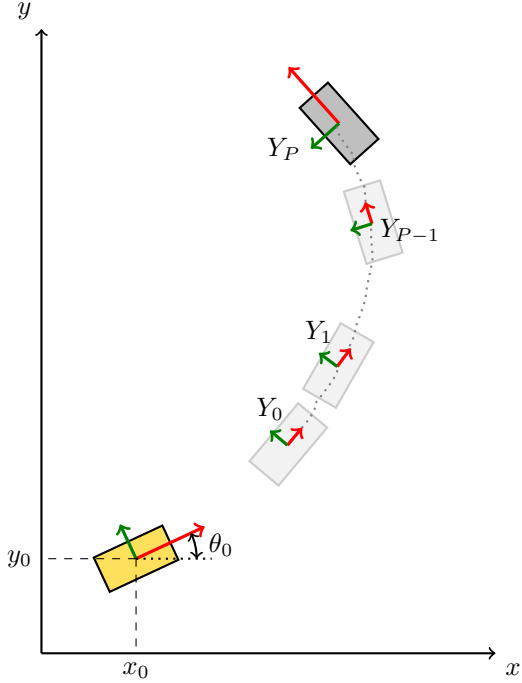
\begin{figure}
    \centering
    \begin{tikzpicture}

    \pgfmathsetmacro{\rectwidth}{1} 
    \pgfmathsetmacro{\rectheight}{0.5} 

    \coordinate (xy_1) at (3.2500,2.7500);
    \coordinate (xy_2) at (3.6119,3.2946);
    \coordinate (xy_3) at (3.9081,3.7897);
    \coordinate (xy_4) at (4.1387,4.2793);
    \coordinate (xy_5) at (4.2866,4.7636);
    \coordinate (xy_6) at (4.3772,5.2320);
    \coordinate (xy_7) at (4.3673,5.6769);
    \coordinate (xy_8) at (4.2867,6.1265);
    \coordinate (xy_9) at (4.1384,6.5666);
    \coordinate (xy_10) at (3.9179,6.9766);

    \pgfmathsetmacro{\thetalA}{50.0000} 
    \pgfmathsetmacro{\thetalB}{53.4287} 
    \pgfmathsetmacro{\thetalC}{60.1650} 
    \pgfmathsetmacro{\thetalD}{69.8055} 
    \pgfmathsetmacro{\thetalE}{81.6362} 
    \pgfmathsetmacro{\thetalF}{94.5766} 
    \pgfmathsetmacro{\thetalG}{107.4181} 
    \pgfmathsetmacro{\thetalH}{118.6476} 
    \pgfmathsetmacro{\thetalI}{127.0589} 
    \pgfmathsetmacro{\thetalJ}{131.8685} 

    \draw[->, thick] (0,0) -- (6,0) node[below right] {$x$};
    \draw[->, thick] (0,0) -- (0,8.25) node[above left] {$y$};

    \pgfmathsetmacro{\thetaf}{25} 
    \pgfmathsetmacro{\xf}{0.75} 
    \pgfmathsetmacro{\yf}{1} 
    \pgfmathsetmacro{\cxf}{\xf + \rectwidth / 2}
    \pgfmathsetmacro{\cyf}{\yf + \rectheight / 2}
    \draw[rotate around={\thetaf:(\cxf,\cyf)},fill=yellow!75!orange!75, thick] (\xf,\yf) 
        rectangle ({\xf + \rectwidth}, {\yf + \rectheight});
    \draw[rotate around={\thetaf:(\cxf,\cyf)},very thick,->,red] (\cxf,\cyf) -- ++(1.0,0);
    \draw[rotate around={\thetaf:(\cxf,\cyf)},very thick,->,green!50!black] (\cxf,\cyf) -- ++(0,0.5);
    \draw[dashed] (\cxf,\cyf) -- (\cxf,0); 
    \draw[dashed] (\cxf,\cyf) -- (0,\cyf); 
    \node[below] at (\cxf,0) {$x_0$}; 
    \node[left] at (0,\cyf) {$y_0$}; 
    \draw[dotted,thick] (\cxf,\cyf) -- ++(1.,0);
    \path[draw,thick, <->] (\cxf,\cyf) ++(0.8,0) arc[start angle=0,end angle=\thetaf,radius=0.8] coordinate (angle_theta_f);
    \node[right of=angle_theta_f, xshift=-0.6cm, yshift=-0.15cm] {$\theta_0$};

    \pgfextractx{\cxl}{\pgfpointanchor{xy_10}{center}}
    \pgfextracty{\cyl}{\pgfpointanchor{xy_10}{center}}   
    \pgfmathsetmacro{\cxl}{\cxl/28.3465}  
    \pgfmathsetmacro{\cyl}{\cyl/28.3465}
    \pgfmathsetmacro{\xl}{\cxl - \rectwidth / 2} 
    \pgfmathsetmacro{\yl}{\cyl - \rectheight / 2} 
    \draw[rotate around={\thetalJ:(\cxl,\cyl)},fill=gray!50, thick] (\xl,\yl) 
        rectangle ({\xl + \rectwidth}, {\yl + \rectheight});
    \draw[rotate around={\thetalJ:(\cxl,\cyl)},very thick,->,red] (\cxl,\cyl) -- ++(1.0,0);
    \draw[rotate around={\thetalJ:(\cxl,\cyl)},very thick,->,green!50!black] (\cxl,\cyl) -- ++(0,0.5) node[black, above right, anchor=east] {$Y_{P}$};

    \pgfextractx{\cxlf}{\pgfpointanchor{xy_1}{center}}
    \pgfextracty{\cylf}{\pgfpointanchor{xy_1}{center}}   
    \pgfmathsetmacro{\cxlf}{\cxlf/28.3465}  
    \pgfmathsetmacro{\cylf}{\cylf/28.3465}
    \pgfmathsetmacro{\xl}{\cxlf - \rectwidth / 2} 
    \pgfmathsetmacro{\yl}{\cylf - \rectheight / 2} 
    \draw[rotate around={\thetalA:(\cxlf,\cylf)}, fill=gray!50, thick, opacity=0.2] (\xl,\yl) 
        rectangle ({\xl + \rectwidth}, {\yl + \rectheight});

    \pgfextractx{\cxls}{\pgfpointanchor{xy_3}{center}}
    \pgfextracty{\cyls}{\pgfpointanchor{xy_3}{center}}   
    \pgfmathsetmacro{\cxls}{\cxls/28.3465}  
    \pgfmathsetmacro{\cyls}{\cyls/28.3465}
    \pgfmathsetmacro{\xl}{\cxls - \rectwidth / 2} 
    \pgfmathsetmacro{\yl}{\cyls - \rectheight / 2} 
    \draw[rotate around={\thetalC:(\cxls,\cyls)}, fill=gray!50, thick, opacity=0.2] (\xl,\yl) 
        rectangle ({\xl + \rectwidth}, {\yl + \rectheight});

    \pgfextractx{\cxlt}{\pgfpointanchor{xy_7}{center}}
    \pgfextracty{\cylt}{\pgfpointanchor{xy_7}{center}}   
    \pgfmathsetmacro{\cxlt}{\cxlt/28.3465}  
    \pgfmathsetmacro{\cylt}{\cylt/28.3465}
    \pgfmathsetmacro{\xl}{\cxlt - \rectwidth / 2} 
    \pgfmathsetmacro{\yl}{\cylt - \rectheight / 2} 
    \draw[rotate around={\thetalG:(\cxlt,\cylt)}, fill=gray!50, thick, opacity=0.2] (\xl,\yl) 
        rectangle ({\xl + \rectwidth}, {\yl + \rectheight});

    \pgfmathsetmacro{\controlDist}{0.1}
    \draw[gray, dotted, thick]
        (xy_1) .. controls +(\thetalA:\controlDist) and +(-\thetalB:\controlDist) .. (xy_2)
        .. controls +(\thetalB:\controlDist) and +(-\thetalC:\controlDist) .. (xy_3)
        .. controls +(\thetalC:\controlDist) and +(-\thetalD:\controlDist) .. (xy_4)
        .. controls +(\thetalD:\controlDist) and +(-\thetalE:\controlDist) .. (xy_5)
        .. controls +(\thetalE:\controlDist) and +(-\thetalF:\controlDist) .. (xy_6)
        .. controls +(\thetalF:\controlDist) and +(-\thetalG:\controlDist) .. (xy_7)
        .. controls +(\thetalG:\controlDist) and +(-\thetalH:\controlDist) .. (xy_8)
        .. controls +(\thetalH:\controlDist) and +(-\thetalI:\controlDist) .. (xy_9)
        .. controls +(\thetalI:\controlDist) and +(-\thetalJ:\controlDist) .. (xy_10);

    \draw[rotate around={\thetalA:(xy_1)},very thick,->,red] (xy_1) -- ++(0.3,0);
    \draw[rotate around={\thetalA:(xy_1)},very thick,->,green!50!black] (xy_1) -- ++(0,0.3) node[black, below left, anchor=south] {$Y_0$};

    \draw[rotate around={\thetalB:(xy_3)},very thick,->,red] (xy_3) -- ++(0.3,0);
    \draw[rotate around={\thetalB:(xy_3)},very thick,->,green!50!black] (xy_3) -- ++(0,0.3) node[black, below left, anchor=south] {$Y_1$};

    \draw[rotate around={\thetalG:(xy_7)},very thick,->,red] (xy_7) -- ++(0.3,0);
    \draw[rotate around={\thetalG:(xy_7)},very thick,->,green!50!black] (xy_7) -- ++(0,0.3) node[xshift=0.25cm, black, below right, anchor=west] {$Y_{P-1}$};

    \end{tikzpicture}

    \caption{Leader (gray) and follower (yellow) on a plane. Solid objects represent current poses, while shaded objects represent previous poses.}
    \label{fig:leader-follower}
\end{figure}

Let the vehicle be modeled as a unicycle with the configuration represented as a vector $\mathbf{q}_t=[x_t,y_t,\theta_t]\in\Real^{3}$. The kinematics of the vehicle are 
\begin{align}\label{eq:SE2}
    x_{t+1} &= x_{t} + v_t \cos(\theta_t) dt \nonumber\\
    y_{t+1} &= y_{t} + v_t \sin(\theta_t) dt \\
    \theta_{t+1} &= \theta_{t} + \omega_t dt \nonumber
\end{align}
where $v_t$ and $\omega_t$ are the linear and angular velocities and $dt$ is the discretization time.

Alternatively, we may state the model on the $\text{SE}(2)$ group. The velocity of the vehicle lies on the tangent space of the $\text{SE}(2)$ group, referred to as the Lie algebra, $\mathfrak{se}(2)$, of the Lie group. The map of a velocity vector $\mathbf{u}$ to the Lie algebra, $(*)^{\wedge}$, is defined as 
\begin{align}\label{eq:SE2}
    (\mathbf{u})^{\wedge} = \begin{bmatrix}
        0 & -\omega & v_{x} \\
        \omega & 0 & v_{y} \\
        0 & 0 & 0
    \end{bmatrix}.
\end{align}
The inverse of the operation is denoted as $(* )^{\vee}$. We may then describe the motion, with a continuous time model,  as
\begin{align}\label{eq:se2}
    \dot{\chi}(t) = \chi(t)\big(\mathbf{u}(t)\big)^{\wedge}
\end{align}
with $\dot{\chi}(t)$ being the speed of the vehicle at time $t$.

In this problem, we consider the control of a follower vehicle that attempts to match an arbitrary trajectory of a leader vehicle with a fixed delay of $P$ timesteps. At each timestep, the follower vehicle receives an observation of the leader relative to itself, its own inputs, and an estimate of the leaders angular and linear velocities
\begin{equation}
    \mathbf{o}_t = [x_t, y_t, \theta_t, \mathbf{u}_t, \mathbf{v}^{L}_{t}] \label{eq:obs}
\end{equation}
with $\mathbf{u}_t$ being the ego vehicle's inputs and $\mathbf{v}^{L}_{t}$ being the leader vehicle linear and angular velocities. In the accompanying image, this would be the relative location $Y_P$ of the leader vehicle in the follower vehicles reference frame.

The goal of the follower is to occupy the same spot as the leader with a fixed-time lag denoted by $P$ timesteps. In \figurename~\ref{fig:leader-follower}, the follower would aim to occupy the pose $Y_0$ at $t=P$, $Y_1$ at $t=(1+P)$, and so on. Practically, this is accomplished by collecting the follower vehicle inputs $\mathbf{u}_{t-P+1:t-1}$ and observations of leader states $\mathbf{o}_{t-P:t}$ in order to generate the predictions of the lagged leader reference poses relative to the current follower vehicle position $\hat{d}_{t:t+P-1}$, where the timesteps are shifted to denote that this is the desired future trajectory of the follower. Note that since the motion of the follower vehicle is uncertain and the vehicle does not have access to global (or local, for that matter) positioning, the uncertainty of future desired locations grows with time. Furthermore, note that in \figurename~\ref{fig:leader-follower}, the indices go from $0$ to $P$ because we are assuming that this configuration refers to the start of the run.

\subsection{Trajectory Estimation Model}

\begin{figure}
    \centering
    \includegraphics[width=0.9\linewidth]{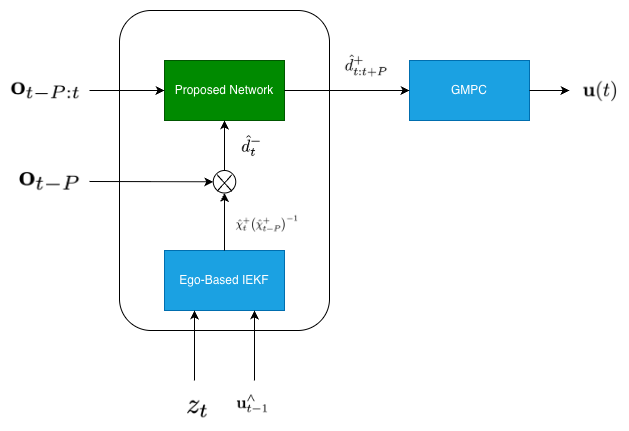}
    \caption{Graphical description of inputs and outputs of the proposed system, with learned components in green and static components in blue.}
    \label{fig:sys-des}
\end{figure}

We propose a novel system for trajectory tracking on a manifold for the time-delayed leader-follower convoy problem described in the previous section. Our system utilizes a Seq2Seq neural network to transform a sequence of \red{relative }observations into a trajectory on a manifold. \red{An IEKF tracks the ego position of the vehicle, and provide an uncorrected prediction of the leaders state a fixed time in the past}. A graphical representation of the proposed system can be seen in \figurename~\ref{fig:sys-des}. Here, we discuss the internal structure of the proposed architecture, and we detail its inputs and outputs in the sequel.

\begin{figure}
    \centering
    \includegraphics[width=0.9\linewidth]{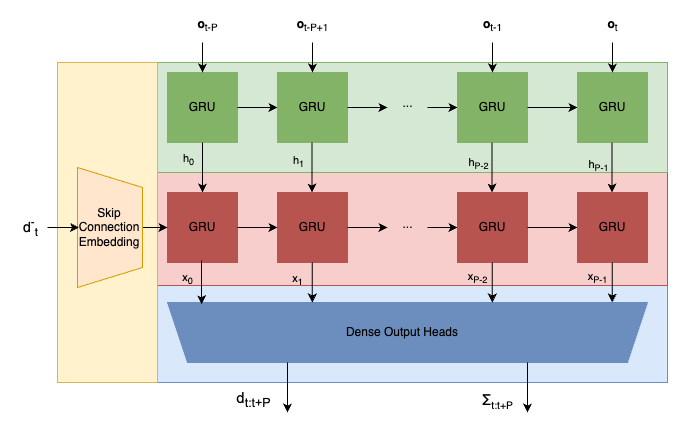}
    \caption{Graphical depiction of the proposed neural network architecture. The green components represent the encoder, the red the decoder, the blue the output modules, and the yellow the skip connection that connects an IEKF estimate through an embedding to the latent space of the decoder.}
    \label{fig:nn-des}
\end{figure}

The Seq2Seq network generates an ego-centered trajectory at every timestep. In addition to the standard structure of the system, we add a connection to the latent space of the model to add an initial estimate of the state to the input, obtained using an IEKF. \red{We note that the IEKF tracks only the ego robot's pose, providing an uncorrected prediction of the initial desired pose. The neural network is then responsible for reconstructing the full trajectory estimates and associated uncertainties.} This structure is shown in \figurename~\ref{fig:nn-des}. \red{An ablation study indicates that the IEKF is not strictly necessary, but improves performance.} 

The neural network is composed of three components: an encoder, a decoder and a probabilistic output module. The encoder is a Gated Recurrent Unit (GRU) layer that receives the last $P$ observations. The hidden values of this layer are passed to the decoder, also composed of a layer of GRUs. The decoder module receives the hidden values of the encoder, and an embedding of the initial estimate of the IEKF, composed of a linear projection to the appropriate dimensionality. Its outputs are passed to the probabilistic output module, composed of a number of feedforward layers, which calculates the means and covariances of the trajectory estimates. The neural network employs the GELU activation function~\cite{gelu}. We provide the code for the neural networks, dataset generation, and evaluation\footnote{https://anonymous.4open.science/r/arbitrary-trajectory-seq2seq-8811/}. 

\subsection{Trajectory Estimation Procedure}

During a forward pass, the Seq2Seq model estimates a trajectory using inputs defined in \eqref{eq:obs} over the last $P$ timesteps, as shown in \figurename~\ref{fig:nn-des}, as well as an estimate of the initial pose obtained using the IEKF. 
The IEKF estimate is
\begin{align}
    \hat{d}^{-}_t = \hat{\chi}^{+}_{t} \big(\hat{\chi}^{+}_{t-P}\big)^{-1}\mathbf{o}_{t-P}
\end{align}
or the transformation of \red{the observation} of the leader at time $t-P$ into the current ego reference frame. This is fed through the skip connection, as shown in the yellow segment on the left of \figurename~\ref{fig:nn-des}.
The network's prediction at timestep $t$ is given by
\begin{equation}
    \hat{d}^{+}_{t:t+P}, \Sigma_{t:t+P} = \hat{f}^{\theta}(\mathbf{o}_{t-P:t}, \hat{d}^{-}_{t})
\end{equation}
where $d^{+}_{t:t+P}$ represents the estimated mean trajectory poses and $\Sigma_{t:t+P}$ the corresponding covariances. In practice, $\mathbf{o}_{t-P:t}$ is given to the initial encoding layer to extract temporal relationships, while $\hat{d}^{-}_t$ is fed through a dense embedding layer before being added as an input to the initial GRU pass in the encoding. This initial estimate is combined with the trajectory encoding layer outputs and passed to the probabilistic output module. The mean output head computes
\begin{equation}
    m(\mathbf{x}) = [\hat{x}, \hat{y}, \sin(\hat{\theta}), \cos(\hat{\theta})]
\end{equation}
ensuring outputs align with \eqref{eq:SE2}. Orientation outputs are clipped to $[-1,1]$ to respect $\sin$ and $\cos$ limits.

The covariance output head, given the input $\mathbf{x}$, is a function $c(\mathbf{x})$ that produces a vector of size $(n+n(n-1)/2)$, with $n$ being the size of the mean vector, in this case, $4$. The components of the vector encode both autocovariances and Pearson correlation coefficients, ensuring valid covariance matrices. The diagonal elements of the covariance matrix are computed as
\begin{align}
    \Sigma_{ii} = \sigma_{i}^{2} = \exp\big(c(\mathbf{x})_{i}\big), i\in n,
\end{align}
while the off-diagonal elements are 
\begin{align}
    \Sigma_{ij} = \rho_{ij} \sigma_i \sigma_j = \tanh\big(c(\mathbf{x})_{n+i}\big)\sqrt{\Sigma_{ii}\Sigma_{jj}}
\end{align}
where $\Sigma_{ij}=\Sigma_{ji}$ for $j < i$. For further details on constructing the $\Sigma$ matrices, see \cite{russel2019}.

\subsection{Training and Losses}

The neural network is trained using a negative log-likelihood~(NLL) loss based on the $\boxminus$ operator~\cite{sun2021idol,russel2019}, adapted for $\text{SE(2)}$. 
The per-timestep loss is defined as
\begin{equation}
    L(d_t, \hat{\chi}_t, \Sigma_t) = \frac{1}{2} (\chi_t \boxminus \hat{\chi}_t ) \Sigma^{-1} (\chi_t \boxminus \hat{\chi}_t ) + \frac{1}{2} ln |\Sigma_t|
\end{equation}
where $\chi_t$ is the ground-truth state at $t$,
$\hat{\chi}_t$ the predicted state, and $\Sigma_t$ the covariance matrix for the prediction.

During training, Gaussian noise is added to the true initial trajectory state to simulate the noisy warm start of the IEKF (yellow block in \figurename~\ref{fig:nn-des}). Experimentally, we find that this improves robustness to out-of-distribution observations compared to using a true IEKF estimate. \red{Through tuning, we find that zero mean input noise with a standard deviation of $0.2\text{m}^2$ performs well. We note that this is a higher standard deviation than the noise sampled in the simulation, detailed in Section \ref{sec:Sims}, implying that this improves robustness to systematic biases.}

\subsection{Dataset Generation}
Training the Seq2Seq model requires a dataset mapping sequences of observations to relative desired trajectories. We use a unicycle kinematic model, though the approach is independent of specific kinematics or dynamics.

Dataset generation involves a vehicle kinematic model, a controller, and a rapidly-exploring random tree~(RRT) path planner. Paths are converted into trajectories via cubic spline interpolation between RRT waypoints.

We first generate a map of random \red{static} obstacles placed in a workspace. This allows the creation of curved paths that avoid obstacles, simulating a leader that avoids real-world features. An RRT planner computes a path from a randomly sampled start point $x_{\text{init}} \in \text{free} $ to a randomly sampled end point $x_{\text{goal}} \in \text{free}$, which is then converted into a trajectory.

The movement of the leader vehicle is simulated by translating future desired trajectory points ($P$ timesteps ahead) into the ego-centered coordinate frame, adding observation noise. Leader inputs are obtained using a proportional controller with exact feedback linearization~\cite{feedback}. These are combined with the ego vehicle measurements, the ego inputs, to form trajectory segments with inputs $\mathbf{o}_{t:t+p}$ and $\hat{d}^-_{t}$, as well as targets $d_{t:t+P}$ corresponding with the NN inputs and outputs in \figurename~\ref{fig:nn-des}, to train the neural network.

To enhance generalization by expanding the input and observation distributions, the ego vehicle is controlled using three different control methods: proportional control with exact feedback linearization, GMPC, and uniform random input sampling. In total, we record $1000$ individual trajectories. 

\subsection{Integration with GMPC}
To optimize tracking, we apply a GMPC method~\cite{gmpc} that incorporates vehicle kinematics on a manifold, leveraging extended trajectory information for smoother control than a proportional–integral–derivative (PID) controller.

Since our system outputs a relative state, we define the tracking error as the negative predicted trajectory:
\begin{align}
    \Psi(t) = -\hat{d}(t) \in \text{SE}(2)
\end{align}
where $\hat{d}(t)$ is the NN's output at time $t$.

To constrain acceleration, we limit input differences between  timesteps. With this constraint, the optimization problem becomes
\begin{eqnarray}
\label{eq:gmpc}
   \min_{u(t)} \boldsymbol{\psi}(T)^{T} Q_{f} \boldsymbol{\psi}(T) + \sum_{t=0}^{T-1} \boldsymbol{\psi}(t)^{T} Q \boldsymbol{\psi}(t) + \mathbf{u}_{k}^{T} H_{k} \mathbf{u}_{k}
\end{eqnarray}
subject to: 
\begin{eqnarray}
\label{eq:gmpc_constraints}
    \boldsymbol{\psi}_{t+1} = A_{t}\boldsymbol{\psi}_{t} + B_{t}\mathbf{u}_{t} + \mathbf{c}_{t}, \nonumber\\
    u_{\text{min}} \leq \mathbf{u}(t) \leq u_{\text{max}}, \nonumber\\
    |\mathbf{u}(t+1) - \mathbf{u}(t)| \leq \delta u_{\text{max}}, 
\end{eqnarray}
with $\delta u_{\text{max}}$ as a constant maximum acceleration and $\psi(0) = -\hat{d}^{+}_{0} \in \text{SE}(2)$, $0 \leq t \leq T-1$.
This reduces transient tracking errors from imperfect Seq2Seq predictions, ensuring smoother motion.

\red{We note that, in the current system, we do not make explicit use of the uncertainty estimate. To date, no robust GMPC variants have yet been proposed. Despite this, an ablative study indicates that the incorporation of probabilistic estimations leads to improved performance, in line with conclusions from representation learning~\cite{LESORT2018379}. By further exploring the development of robust on-manifold MPC, we plan to extend our work in the future to improve constraint satisfaction and stability by incorporating our uncertainty estimates into the controller.}

\section{Numerical Examples}\label{sec:Sims}

We evaluate the proposed system through simulation. Performance is benchmarked against an \red{IEKF-based classical estimation technique instantiated with the exact process and measurement noise}, \red{a standard GRU-based RNN approach, and a transformer model, both trained using MSE-loss}.
For state-feedback evaluation, we compare against the true trajectory to establish a lower bound on tracking error. \red{Unless stated otherwise, all measurements in simulation have zero-mean Gaussian noise added, with $\sigma=0.14$m, providing lower accuracy than popular mid-range LiDARs~\cite{ouster_os1_datasheet}. Similarly, zero-mean Gaussian noise is added to the inputs of the simulated vehicle, equating to standard deviations of $0.14$m/s and $0.14\theta$/s for the linear and angular velocity, respectively.}

We consider a unicycle model with parameters: $v_{max}=3$, $v_{min}=-3$, $\omega_{max}=3$, $\omega_{min}=-3$, and $dt=0.1$. All vehicles start at rest in the same position. The ego vehicle remains stationary for $P$ timesteps before following the leader's trajectory with a $P$-step delay.

Performance is evaluated using both the standard Euclidean root mean-squared error~(RMSE) and a $\boxminus$-RMSE that accounts for orientation errors:
\begin{equation}
    \boxminus \text{-} RMSE(y, \hat{y}) = \sqrt{\sum_{i=1}^{N} \frac{(\hat{y}_{i} \boxminus y_{i})^{2}}{N}}
\end{equation}
where $y_i$ and $\hat{y}_i$ are the true and predicted values, respectively, and $N$ is the number of samples.
The NN is trained using the parameters in \tablename~\ref{tab:nn_param}.

\begin{table}
    \centering
    \caption{Parameters of the Seq2Seq Model}
    \begin{tabular}{c|c}
        Parameter & Value \\
        \hline
        Learning Rate & $0.005$ \\
        \hline
        Learning Rate Decay &  $0.995$\\
        \hline
        \red{Encoder Layers} & $1$ \\
        \hline
        \red{Decoder Layers} & $1$ \\ 
        \hline
        \red{Encoder Hidden State Dimensionality} & $64$ \\
        \hline
        \red{Decoder Hidden State Dimensionality} & $64$ \\
        \hline
        \red{Output Head Hidden Layer(s) Dimension} & $[64]$ \\
        \hline
        Dropout & $0.1$ \\
        \hline
        Prediction Horizon~($P$) & $100$ \\
        \hline
        L2 Regularization &  $0.0001$ \\
        \hline
        Training Input Noise & $\mathcal{N}(\mathbf{0}, \mathbf{0.2}\text{m})$
    \end{tabular}
    \label{tab:nn_param}
\end{table}

\begin{table}[]
    \centering
    \caption{\red{Dataset Generation Parameter}s}
    \begin{tabular}{c|c}
        \red{Workspace Size} & $200\text{m} \times 200\text{m}$ \\
        \hline
        \red{Number of Obstacles} & $20$  \\
        \hline
        \red{Obstacle Size} & $\mathcal{N}(7 \text{m}, 2 \text{m}^2)$ \\
        \hline
        \red{RRT Number of Iterations} & $400$  \\
        \hline
        \red{RRT Max. Curvature} & $1.0$ \\
    \end{tabular}

    \label{tab:train_vals}
\end{table}

\subsection{System Identification}

We evaluate the trajectory models on a dataset of $50$ trajectories not seen during training. The dataset consists of trajectories with a ratio of $40\%$ feedback linearization~(FL) controllers, $40\%$ GMPC controllers, and $20\%$ random inputs. Results summarized in \tablename~\ref{tab:traj-err} show that the Seq2Seq model achieves a $\boxminus$-RMSE that is approximately $62\%$ of the IEKF, \red{and outperforms both the bidirectional GRU network and the Transformer network}.

\begin{table}
    \centering
    \caption{Prediction Errors of Trajectory Tracking Models}
    \begin{tabular}{c|c|c}
        Method & RMSE & $\boxminus$-RMSE \\
        \hline
        Proposed & $\mathbf{0.569}$ & $\mathbf{0.523}$ \\
        \hline
        IEKF & $0.855$ & $0.846$  \\ 
        \hline
        \red{Bidirectional-GRU} & $0.592$ & $0.580$ \\
        \hline
        \red{Transformer} & $0.653$ & $0.644$ \\
    \end{tabular}
    \label{tab:traj-err}
\end{table}


\subsection{Closed-Loop Performance}



\red{In closed loop simulations, the IEKF and proposed methods use the GMPC while a nonlinear MPC is used with the RNN and transformer models, as they do not make predictions on the $SE(2)$ manifold.} We generate $100$ trajectories distinct from the training data, each $150$ seconds long, and evaluate each system configuration. Results are displayed in \tablename~\ref{tab:closed-loop}.

\begin{table}
    \centering
    \caption{Closed Loop Performance of Trajectory-Models}
    \begin{tabular}{c|c|c}
        Method & RMSE & $\boxminus$-RMSE\\
        \hline
        True Traj. + GMPC & $0.213$  & $0.211$ \\
        \hline
        \red{Proposed} & \red{$0.799$} & \red{$0.649$} \\
        \hline
        IEKF + GMPC & $1.035$ & $0.706$ \\
        \hline
        \red{Bidirectional-GRU + MPC} & $1.502$ & $1.347$\\
        \hline
        \red{Transformer + MPC} & $1.536$ & $1.388$ \\
        
    \end{tabular}
    \label{tab:closed-loop}
\end{table}

The Seq2Seq model's GMPC tracking error is approximately $3$ times higher than the noise-free lower bound. It outperforms the IEKF method on both metrics, while reducing expert design requirements. The difference between the Euclidean RMSE and the $\boxminus$-RMSE is smaller for the Seq2Seq model than the IEKF model, indicating improved estimation of orientation. \red{Both the IEKF and proposed architecture outperform the other learning-based baselines, indicating that the proposed structure is necessary to maintain the integrity of the trajectory estimations when utilizing neural networks.} 

\begin{figure}
    \centering
    \includegraphics[width=0.9\linewidth]{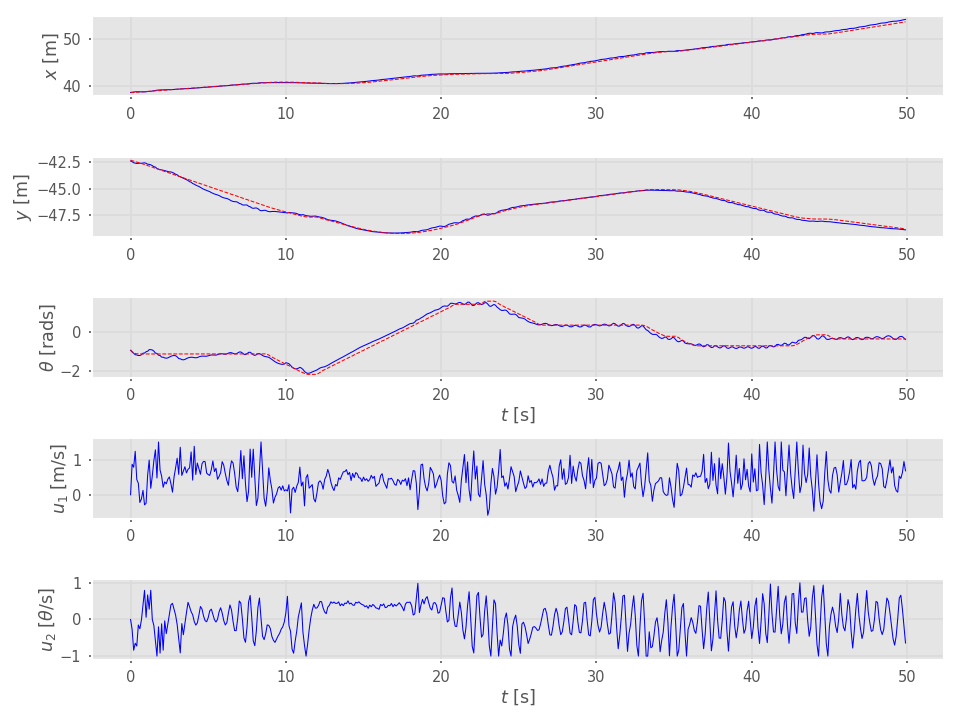}
    \caption{True vehicle states~(blue), desired vehicle states~(red) and control effort for the GMPC-Seq2Seq model system.}
    \label{fig:closed-loop-nn}
\end{figure}

\figurename~\ref{fig:closed-loop-nn} shows the desired trajectory, state history, and control effort for the Seq2Seq model with GMPC. The results were shifted by the predefined fixed delay to compute the errors. Aside from some transient errors in position, the majority of the error is concentrated in the orientation of the ego vehicle. 

\subsection{Ablation Study}
To evaluate the effect of each system component, an ablative study is conducted. We perform four ablations: (i) removing the  skip connection, (ii) replacing the proposed $\boxminus$ loss with an $\mathbb{R}^3$ MSE loss, (iii) applying both modifications, \red{and (iv) utilizing purely predictions from a motion model to estimate the initial desired pose in place of the IEKF}. The resulting trajectory reconstruction errors are reported in Table \ref{tab:ablation-traj-err}.

\begin{table}
    \centering
    \caption{Prediction Error Ablative Study}
    \begin{tabular}{c|c|c}
        Method & RMSE & $\boxminus$-RMSE \\
        \hline
        Proposed Architecture & $\mathbf{0.569}$ & $\mathbf{0.523}$ \\
        \hline
        Proposed Architecture Seq2Seq$+ \mathbb{R}^3$-MSE loss & $0.829$ & $0.822$ \\
        \hline
        Seq2Seq$+\boxminus$-loss & $ 0.683 $ & $ 0.630 $ \\
        \hline
        Seq2Seq $+$ MSE loss & $ 0.889 $ & $ 0.881 $ \\
    \end{tabular}
    \label{tab:ablation-traj-err}
\end{table}

These results indicate that the $\boxminus$ NLL loss has the highest impact \red{o}n the system's ability to reconstruct trajectories. To further assess each component, we evaluate the effects of the ablations on path-tracking performance. For this evaluation, $100$ randomly generated trajectories of $150$ seconds each are used. We utilize GMPC for the $\boxminus$-loss models, and a nonlinear MPC for the $\mathbb{R}^3$-loss models. The results of this ablation are summarized in Table \ref{tab:closed-loop-ablation}.

\begin{table}
    \centering
    \caption{Closed Loop Performance Ablative Study}
    \begin{tabular}{c|c|c}
        Method & RMSE & $\boxminus$-RMSE\\
        \hline
        Proposed & \red{$\mathbf{0.799}$} & \red{$\mathbf{0.649}$} \\
        \hline
        \red{Proposed + Pure Prediction} & \red{$0.794$} & \red{$0.675$} \\
        \hline
        Proposed Architecture$+ \mathbb{R}$-MSE loss & $1.156$ & $1.056$ \\
        \hline
        Seq2Seq$+\boxminus$-loss & $1.766$ & $1.629$ \\
        \hline
        Seq2Seq$+ \mathbb{R}$-MSE loss & $1.460$ & $1.333$ \\

    \end{tabular}
    \label{tab:closed-loop-ablation}
\end{table}

The results indicate that the proposed \red{initial prediction} is crucial to performance. Without it, the $\boxminus$-loss Seq2Seq network performs worse than the Seq2Seq-MSE baseline. Together with the trajectory errors in Table \ref{tab:ablation-traj-err}, this suggests that the $\boxminus$ improves prediction accuracy, \red{but} degrades early-trajectory reconstruction. The addition of the IEKF prediction helps orient the NN, reducing this effect. \red{However, relying only on odometry, rather than an ego-position IEKF, still provides better performance than utilizing the purely classical method. This implies that the neural network is able to synthesize the base prediction with the observations to create more precise trajectories, and that the system can match the performance of a traditional estimation method with no need for corrections from external measurements.}

\subsection{\red{Robustness to Failure Modes}}\label{sec:failure_modes}

\red{Here, we evaluate the closed loop system's robustness to some common failure modes, such as external disturbances and non-Gaussian Noise. We use numerical simulations to perform these evaluations as they provide precise control over the their occurrences.}

\red{To model physical challenges common in real robot scenarios, we inject disturbances into the system. We use two types of disturbances; a globally constant disturbance, and biases on angular velocities~\cite{Prado2003}. We use a value of $0.1$ m/s in in a fixed, arbitrary direction for the constant disturbance, and $0.5\text{ }\theta$/s for the angular velocity bias. Additionally, we vary the the noise applied to the sensor readings, using uniform noise and anisotropic noise. Uniform noise is sampled on the range of $[-0.2,0.2]$ for both the range and heading measurements, and anisotropic noise is modeled by scaling the lateral component of every measurement noise sample by $3$.}

\red{We summarize the evaluation in Table \ref{tab:failure_modes}. The system performance is robust to most variations in noise and disturbances. When considering constant disturbances, we observe a larger loss of performance. As the proposed system relies primarily on relative observations, shifts that do not correspond with nominal system dynamics are difficult to account for. However, we still maintain sub-meter tracking errors.}

\begin{table}[]
    \centering
    \caption{\red{Summary Of Performance Under Introduced Failure Modes For $20$ Trajectories.}}
    \label{tab:failure_modes}
    \begin{tabular}{c|c|c}
        \red{\textbf{Failure Mode}} & \red{\textbf{RMSE}} & \red{\textbf{$\boxminus$-RMSE}} \\ \hline
        \red{Nominal Conditions} & \red{$0.743$}  & \red{$0.629$} \\ \hline
        \red{Uniform Noise} & \red{$0.746$} &  \red{$0.632$}  \\ \hline
        \red{Anisotropic Noise} & \red{$0.750$} & \red{$0.638$} \\ \hline
        \red{Constant Disturbances} & \red{$0.944$} & \red{$0.820$} \\ \hline
        \red{Angular Velocity Bias} & \red{$0.741$} & \red{$0.632$} \\ \hline 
    \end{tabular}
\end{table}

\section{Experiments} \label{sec:exp}
\begin{figure}
    \centering
    \includegraphics[width=0.9\linewidth]{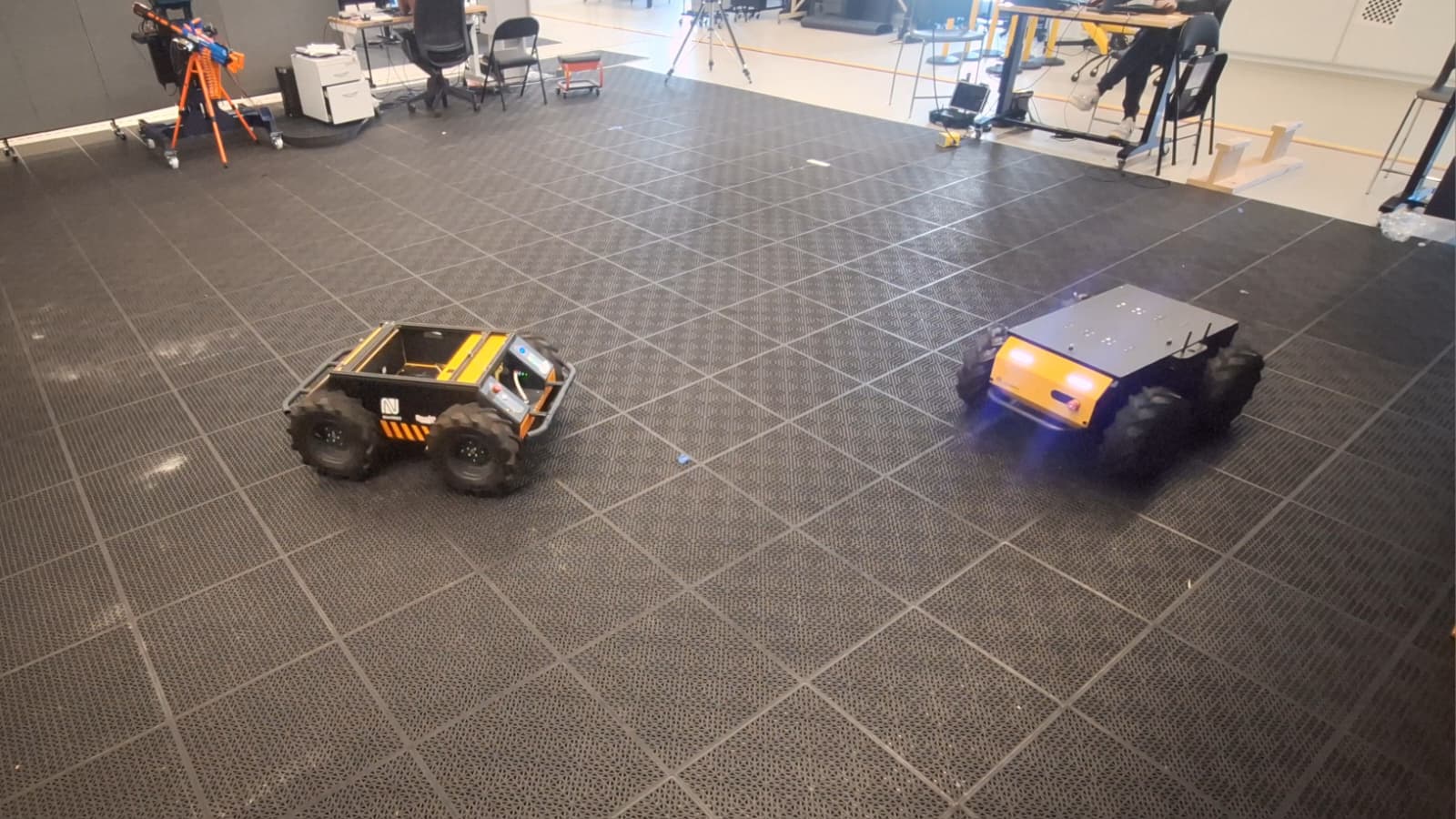}
    \caption{Experiment setup, with the leader (left) and follower (right).}
    \label{fig:robots}
\end{figure}

To validate our method, we deploy our convoy system on real robots in a controlled indoor environment. \red{We note that, despite being known to the operators \textit{a priori}, the robots have no map of the environment.} Experiments are conducted using Clearpath Husky differential drive robots with ROS2 and a Vicon motion capture system. An image of the experiment setup can be seen in \figurename~\ref{fig:robots}. Videos of the experiments are available \hyperlink{here}{https://www.youtube.com/watch?v=Reot4X9-ktE}.

The leader robot is controlled manually, with arbitrary trajectories decided by a human operator. The Vicon system captures the positions of each robot and the velocities of the leader. These are translated into relative distances of the leader vehicle to the follower, which are then passed \red{to} the trajectory reconstruction system. We initialize each experiment with the leader directly in front of the follower. Upon beginning the experiment, the leader is driven by the operator. The follower delays for $10$ seconds, and begins following the trajectory of the leader. This continues until the trajectory is completed by the follower, or the experiment is manually terminated. This is repeated $8$ times. We conduct an additional experiment where the leader is allowed to move more quickly than the follower, thus giving the follower an infeasible path. This allows us to test the robustness of the system to unexpected scenarios.

\begin{figure}
    \centering
    \includegraphics[width=0.9\linewidth]{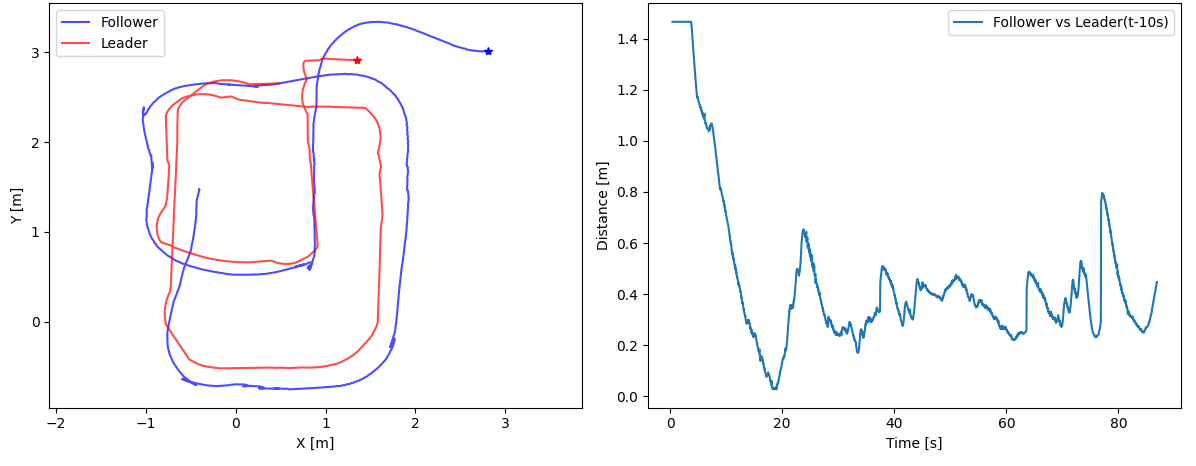}
    \caption{A leader's and follower's trajectories (left) and the separation between the leader's position and the follower's position $10$ seconds later (right).}
    \label{fig:real_robot}
\end{figure}

All observational data has Gaussian noise (matching that used in simulation) added to the Vicon observations.
To ensure path feasibility, the leader's velocity is limited to  $0.3$ m/s and the follower's to $0.4$ m/s. The most complex trajectory along with its position error are shown in Fig.~\ref{fig:real_robot}.

We report the data from the experiments in Table \ref{tab:a}. Overall, the real-world RMSE is higher than in simulation, due to the Sim2Real gap. \red{We note that the mean difference across the experiments in position between the true robot state and the state predicted using the nominal model is $68$cm, indicating unmodelled system dynamics present in the real robot. These disturbances and system performance, are consistent with the findings of the failure mode analysis in Section \ref{sec:failure_modes}.} This discrepancy could be mitigated through additional real-world data collection and transfer learning~\cite{9308468}. Nonetheless, the results indicate that the proposed method remains effective in real robotic deployments.

\begin{table}
    \centering
    \caption{Summary of $9$ Experiments and their associated $\boxminus$-RMSEs.}\label{tab:a}
    \begin{tabular}{c|c}
        Trial & $\boxminus$-RMSE \\
        \hline
        Max. RMSE & $1.005$ \\\hline
        Min. RMSE & $0.722$\\\hline
        Mean & $0.876$ \\\hline
        Edge Case & $0.979$
    \end{tabular}
\end{table}

In addition to the previously noted experiments, we evaluate a trajectory in which the leader vehicle moves faster than the follower, listed in Table \ref{tab:a} as the ``Edge Case". The follower achieves a $\boxminus$-RMSE of $0.979$m, indicating that the system can withstand paths that are infeasible at the cost of higher tracking errors, showing robustness to operational problems.

\section{Conclusion}
In this paper, we proposed and evaluated a novel deep-learning approach for no-communication leader trajectory estimation for a leader-follower convoy on the $\text{SE}(2)$ manifold, \red{integrated with an manifold-based MPC}. The method was tested on a simulated unicycle-model with arbitrary nonlinear trajectories. Further, we validated the method using real-world experiments. The system \red{outperformed both classical and learning baselines, and functioned with minor degradation in real experiments despite model mismatch}. \red{Ablation studies confirm the inclusion of all components of the system}. \red{Real} experiments further show that the follower vehicle \red{is robust to} non-kinematically feasible \red{leader trajectories}.

\subsubsection{\red{Future Work}}
As the proposed model outputs covariance matrices, the use of an uncertainty weighted GMPC cost could improve the tracking error of the method with a small amount of additional computation. \red{The integration of robust manifold-based MPC methods as they become available can aid constraint satisfaction}. Integration with template or model-based perception methods, \red{as well as obstacle avoidance methods,} could improve the applicability of the method to real systems.



%
\balance
\bibliographystyle{IEEEtran}
\bibliography{IEEEabrv,cite}

@article{HERTZBERG201357,
title = {Integrating generic sensor fusion algorithms with sound state representations through encapsulation of manifolds},
journal = {Information Fusion},
volume = {14},
number = {1},
pages = {57-77},
year = {2013},
issn = {1566-2535},
doi = {https://doi.org/10.1016/j.inffus.2011.08.003},
author = {Christoph Hertzberg and René Wagner and Udo Frese and Lutz Schröder}
}

@ARTICLE{gmpc,
  author={Tang, Jiawei and Wu, Shuang and Lan, Bo and Dong, Yahui and Jin, Yuqiang and Tian, Guangjian and Zhang, Wen-An and Shi, Ling},
  journal={IEEE Robotics and Automation Letters}, 
  title={{GMPC: Geometric Model Predictive Control for Wheeled Mobile Robot Trajectory Tracking}}, 
  year={2024},
  volume={9},
  number={5},
  pages={4822-4829},
  doi={10.1109/LRA.2024.3381088}}

@ARTICLE{seq2seqsurv,
  author={Keneshloo, Yaser and Shi, Tian and Ramakrishnan, Naren and Reddy, Chandan K.},
  journal={IEEE Transactions on Neural Networks and Learning Systems}, 
  title={Deep Reinforcement Learning for Sequence-to-Sequence Models}, 
  year={2020},
  volume={31},
  number={7},
  pages={2469-2489},
  doi={10.1109/TNNLS.2019.2929141}}

@INPROCEEDINGS{seq2seqfore,
  author={Du, Shengdong and Li, Tianrui and Horng, Shi-Jinn},
  booktitle={2018 9th International Symposium on Parallel Architectures, Algorithms and Programming (PAAP)}, 
  title={Time Series Forecasting Using Sequence-to-Sequence Deep Learning Framework}, 
  year={2018},
  volume={},
  number={},
  pages={171-176},
  doi={10.1109/PAAP.2018.00037}}

@inproceedings{sun2021idol,
  title={IDOL: Inertial deep orientation-estimation and localization},
  author={Sun, Scott and Melamed, Dennis and Kitani, Kris},
  booktitle={Proceedings of the AAAI Conference on Artificial Intelligence},
  volume={35},
  number={7},
  pages={6128--6137},
  year={2021}
}

@ARTICLE{russel2019,
  author={Russell, Rebecca L. and Reale, Christopher},
  journal={IEEE Transactions on Neural Networks and Learning Systems}, 
  title={Multivariate Uncertainty in Deep Learning}, 
  year={2022},
  volume={33},
  number={12},
  pages={7937-7943},
  doi={10.1109/TNNLS.2021.3086757}}

@ARTICLE{iekf,
  author={Barrau, Axel and Bonnabel, Silvère},
  journal={IEEE Transactions on Automatic Control}, 
  title={The Invariant Extended Kalman Filter as a Stable Observer}, 
  year={2017},
  volume={62},
  number={4},
  pages={1797-1812},
  doi={10.1109/TAC.2016.2594085}}

@article{nahavandi_autonomous_2022,
	title = {Autonomous {Convoying}: {A} {Survey} on {Current} {Research} and {Development}},
	volume = {10},
	copyright = {https://creativecommons.org/licenses/by/4.0/legalcode},
	issn = {2169-3536},
	shorttitle = {Autonomous {Convoying}},
	doi = {10.1109/ACCESS.2022.3147251},
	language = {en},
	journal = {IEEE Access},
	author = {Nahavandi, Saeid and Mohamed, Shady and Hossain, Ibrahim and Nahavandi, Darius and Salaken, Syed Moshfeq and Rokonuzzaman, Mohammad and Ayoub, Rachael and Smith, Robin},
	year = {2022},
	pages = {13663--13683},
}

@INPROCEEDINGS{ZhaoConvoy,
  author={Zhao, Xijun and Yao, Wen and Li, Ning and Wang, Yang},
  booktitle={2017 IEEE International Conference on Unmanned Systems (ICUS)}, 
  title={Design of leader's path following system for multi-vehicle autonomous convoy}, 
  year={2017},
  volume={},
  number={},
  pages={132-138},
  doi={10.1109/ICUS.2017.8278329}}

@article{petrov_nonlinear_nodate,
	author = {Petrov, Plamen},
	title = {Nonlinear {Adaptive} {Control} of a {Two}-{Vehicle} {Convoy}},
	language = {en},
      journal={The Open Cybernetics \& Systemics Journal},
  volume={3},
  number={},
  pages={70--78},
  year={2009}
   
}

@article{petrov_mathematical_nodate,
author = {Petrov, Plamen},
year = {2008},
month = {01},
journal = {WSEAS TRANSACTIONS on SYSTEMS and CONTROL},
pages = {835-848},
title = {A Mathematical Model for Control of an Autonomous Vehicle Convoy},
volume = {3}
}

@INPROCEEDINGS{monkeySee,
  author={Wang, Kai and Givigi, Sidney and Marshall, Joshua A.},
  booktitle={2024 IEEE International Systems Conference (SysCon)}, 
  title={Monkey See, Monkey Do: Constant Time Delay Leader Following for Wheeled Mobile Robots using Uncertainty-Tuned Model Predictive Control}, 
  year={2024},
  volume={},
  number={},
  pages={1-8},
  doi={10.1109/SysCon61195.2024.10553558}}

@ARTICLE{feedback,
  author={Boukas, T.K. and Habetler, T.G.},
  journal={IEEE Transactions on Power Electronics}, 
  title={High-performance induction motor speed control using exact feedback linearization with state and state derivative feedback}, 
  year={2004},
  volume={19},
  number={4},
  pages={1022-1028},
  doi={10.1109/TPEL.2004.830042}}

@INPROCEEDINGS{Fries,
  author={Fries, Carsten and Luettel, Thorsten and Wuensche, Hans-Joachim},
  booktitle={2013 IEEE Intelligent Vehicles Symposium (IV)}, 
  title={Combining model- and template-based vehicle tracking for autonomous convoy driving}, 
  year={2013},
  volume={},
  number={},
  pages={1022-1027},
  doi={10.1109/IVS.2013.6629600}}

@Article{Yaacoub2022,
author={Yaacoub, Jean-Paul A.
and Noura, Hassan N.
and Salman, Ola
and Chehab, Ali},
title={Robotics cyber security: vulnerabilities, attacks, countermeasures, and recommendations},
journal={International Journal of Information Security},
year={2022},
month={Feb},
day={01},
volume={21},
number={1},
pages={115-158},
issn={1615-5270},
doi={10.1007/s10207-021-00545-8},
}

@INPROCEEDINGS{WuConsensus,
  author={Wu, Shuang and Xia, Yuanqing and Lin, Min and Luo, Yu},
  booktitle={2018 Chinese Automation Congress (CAC)}, 
  title={Leader-following Consensus and Trajectory Tracking for Nonholonomic Mobile Robots}, 
  year={2018},
  volume={},
  number={},
  pages={3678-3683},
  doi={10.1109/CAC.2018.8623591}}

@INPROCEEDINGS{SantosFuzzy,
  author={Santos, Carlos and Espinosa, Felipe and Pizarro, Daniel and Valdés, Fernando and Santiso, Enrique and Díaz, Isabel},
  booktitle={2010 IEEE 15th Conference on Emerging Technologies \& Factory Automation (ETFA 2010)}, 
  title={Fuzzy Decentralized Control for guidance of a convoy of robots in non-linear trajectories}, 
  year={2010},
  volume={},
  number={},
  pages={1-8},
  doi={10.1109/ETFA.2010.5641175}}

@INPROCEEDINGS{WangVision,
  author={Wang, Yaqin and Stanković, Miloš and Smith, Anthony and Matson, Eric T},
  booktitle={2021 IEEE Sensors Applications Symposium (SAS)}, 
  title={Leader-Follower System in Convoys:: An Experimental Design Focusing on Computer Vision}, 
  year={2021},
  volume={},
  number={},
  pages={1-6},
  doi={10.1109/SAS51076.2021.9530146}}

@article{zhang_leader-follower_2024,
	title = {Leader-{Follower} {Formation} {Control} of {Perturbed} {Nonholonomic} {Agents} {Along} {Parametric} {Curves} {With} {Directed} {Communication}},
	volume = {9},
	copyright = {https://ieeexplore.ieee.org/Xplorehelp/downloads/license-information/IEEE.html},
	issn = {2377-3766, 2377-3774},
	doi = {10.1109/LRA.2024.3445657},
	language = {en},
	number = {10},
	journal = {IEEE Robotics and Automation Letters},
	author = {Zhang, Bin and Shao, Xiaodong and Zhi, Hui and Qiu, Liuming and Romero, Jose Guadalupe and Navarro-Alarcon, David},
	month = oct,
	year = {2024},
	pages = {8603--8610},
}

@article{walter_uvdar_2019,
	title = {{UVDAR} {System} for {Visual} {Relative} {Localization} {With} {Application} to {Leader}–{Follower} {Formations} of {Multirotor} {UAVs}},
	volume = {4},
	copyright = {https://ieeexplore.ieee.org/Xplorehelp/downloads/license-information/IEEE.html},
	issn = {2377-3766, 2377-3774},
	doi = {10.1109/LRA.2019.2901683},
	language = {en},
	number = {3},
	journal = {IEEE Robotics and Automation Letters},
	author = {Walter, Viktor and Staub, Nicolas and Franchi, Antonio and Saska, Martin},
	month = jul,
	year = {2019},
	pages = {2637--2644},
}

@article{han_leader-follower_2024,
	title = {The {Leader}-{Follower} {Formation} {Control} of {Nonholonomic} {Vehicle} {With} {Follower}- {Stabilizing} {Strategy}},
	volume = {9},
	copyright = {https://ieeexplore.ieee.org/Xplorehelp/downloads/license-information/IEEE.html},
	issn = {2377-3766, 2377-3774},
	doi = {10.1109/LRA.2023.3307010},
	language = {en},
	number = {1},
	journal = {IEEE Robotics and Automation Letters},
	author = {Han, Seungho and Yang, Seunghoon and Lee, Yeongseok and Lee, Minyoung and Park, Ji-il and Kim, Kyung-Soo},
	month = jan,
	year = {2024},
	pages = {707--714},
}

@article{denasi_independent_2018,
	title = {Independent and {Leader}–{Follower} {Control} for {Two} {Magnetic} {Micro}-{Agents}},
	volume = {3},
	copyright = {https://ieeexplore.ieee.org/Xplorehelp/downloads/license-information/IEEE.html},
	issn = {2377-3766, 2377-3774},
	doi = {10.1109/LRA.2017.2737484},
	language = {en},
	number = {1},
	journal = {IEEE Robotics and Automation Letters},
	author = {Denasi, Alper and Misra, Sarthak},
	month = jan,
	year = {2018},
	pages = {218--225},
}

@article{gelu,
  author       = {Dan Hendrycks and
                  Kevin Gimpel},
  title        = {Bridging Nonlinearities and Stochastic Regularizers with Gaussian
                  Error Linear Units},
  journal      = {CoRR},
  volume       = {abs/1606.08415},
  year         = {2016},
  eprinttype    = {arXiv},
  eprint       = {1606.08415},
  bibsource    = {dblp computer science bibliography, https://dblp.org}
}

@Article{cheung,
AUTHOR = {Cheung, Calvin and Rawashdeh, Samir and Mohammadi, Alireza},
TITLE = {Jam Mitigation for Autonomous Convoys via Behavior-Based Robotics},
JOURNAL = {Applied Sciences},
VOLUME = {12},
YEAR = {2022},
NUMBER = {19},
ARTICLE-NUMBER = {9863},
ISSN = {2076-3417},
DOI = {10.3390/app12199863}
}

@article{shrivastava_deep_2021,
	title = {A deep learning based approach for trajectory estimation using geographically clustered data},
	volume = {3},
	issn = {2523-3971},
	doi = {10.1007/s42452-021-04556-x},
	number = {6},
	journal = {SN Applied Sciences},
	author = {Shrivastava, Aditya and Verma, Jai Prakash V. and Jain, Swati and Garg, Sanjay},
	month = may,
	year = {2021},
	pages = {597},
}

@article{ZAMBONI2022108252,
title = {Pedestrian trajectory prediction with convolutional neural networks},
journal = {Pattern Recognition},
volume = {121},
pages = {108252},
year = {2022},
issn = {0031-3203},
doi = {https://doi.org/10.1016/j.patcog.2021.108252},
author = {Simone Zamboni and Zekarias Tilahun Kefato and Sarunas Girdzijauskas and Christoffer Norén and Laura {Dal Col}},
}

@INPROCEEDINGS{Park2018,
  author={Park, Seong Hyeon and Kim, ByeongDo and Kang, Chang Mook and Chung, Chung Choo and Choi, Jun Won},
  booktitle={2018 IEEE Intelligent Vehicles Symposium (IV)}, 
  title={Sequence-to-Sequence Prediction of Vehicle Trajectory via LSTM Encoder-Decoder Architecture}, 
  year={2018},
  volume={},
  number={},
  pages={1672-1678},
  doi={10.1109/IVS.2018.8500658}}

@Article{Zhang2013A,
AUTHOR = {Zhang, Yuanben and Han, Zhonghe and Zhou, Xue and Zhang, Lili and Wang, Lei and Zhen, Enqiang and Wang, Sijun and Zhao, Zhihao and Guo, Zhi},
TITLE = {PESO: A Seq2Seq-Based Vessel Trajectory Prediction Method with Parallel Encoders and Ship-Oriented Decoder},
JOURNAL = {Applied Sciences},
VOLUME = {13},
YEAR = {2023},
NUMBER = {7},
ARTICLE-NUMBER = {4307},
ISSN = {2076-3417},
DOI = {10.3390/app13074307}
}

@INPROCEEDINGS{9308468,
  author={Zhao, Wenshuai and Queralta, Jorge Peña and Westerlund, Tomi},
  booktitle={2020 IEEE Symposium Series on Computational Intelligence (SSCI)}, 
  title={Sim-to-Real Transfer in Deep Reinforcement Learning for Robotics: a Survey}, 
  year={2020},
  volume={},
  number={},
  pages={737-744},
  doi={10.1109/SSCI47803.2020.9308468}}

@datasheet{ouster_os1_datasheet,
  title        = {OS1 Mid-Range High-Resolution Imaging Lidar},
  author       = {{Ouster, Inc.}},
  year         = {2022},
  url          = {https://data.ouster.io/downloads/datasheets/datasheet-rev7-v3p0-os1.pdf},
  note         = {Accessed: 2026-02-16}
}

@article{LESORT2018379,
title = {State representation learning for control: An overview},
journal = {Neural Networks},
volume = {108},
pages = {379-392},
year = {2018},
issn = {0893-6080},
doi = {https://doi.org/10.1016/j.neunet.2018.07.006},
url = {https://www.sciencedirect.com/science/article/pii/S0893608018302053},
author = {Timothée Lesort and Natalia Díaz-Rodríguez and Jean-Frano̧is Goudou and David Filliat},
}

@article{Prado2003,
title = {Effects of terrain irregularities on wheeled mobile robot}, 
volume={21},
DOI={10.1017/S0263574702004563},
number={2}, 
journal={Robotica},
author={Prado, Maria and Simón, Antonio and Pérez, Ana and Ezquerro, Francisco}, year={2003}, 
pages={143–152}}

%








\end{document}